\documentclass[runningheads]{llncs}
\usepackage[hidelinks]{hyperref} 
\usepackage[T1]{fontenc}
\usepackage{cite}
\usepackage{amsmath,amssymb,amsfonts}
\usepackage{algorithmic}
\usepackage{graphicx}
\usepackage{textcomp}
\usepackage{orcidlink} 
\usepackage{subcaption}
\usepackage{multirow}
\usepackage{array}
\usepackage{listings}
\usepackage{tabularx}
\usepackage{indentfirst}
\usepackage{color}
\usepackage{hyperref}
\usepackage{tabularray}

\usepackage{orcidlink}
\usepackage{enumitem}
\setlist{nosep}
\newcommand\comment[1]{{\color{blue}#1}}

\usepackage[normalem]{ulem}  
\usepackage{xcolor}          

\newcommand{\repthanks}[1]{\textsuperscript{\ref{#1}}}
\makeatletter
\patchcmd{\maketitle}
{\def\thanks}
{\let\repthanks\repthanksunskip\def\thanks}
{}{}
\patchcmd{\@maketitle}
{\def\thanks}
{\let\repthanks\@gobble\def\thanks}
{}{}
\newcommand\repthanksunskip[1]{\unskip{}}
\makeatother

\usepackage{graphicx}
\begin{document}
\title{VisChronos: Revolutionizing Image Captioning Through Real-Life Events}
%
%
\author{Phuc-Tan Nguyen\thanks{Both authors contributed equally to this research.\protect\label{X}}\inst{1,2}\orcidlink{0009-0004-0720-4038}\and
Hieu Nguyen\repthanks{X}\inst{1,2}\orcidlink{0009-0001-7726-5301} \and
Minh-Triet Tran\inst{1,2}\orcidlink{0000-0003-3046-3041}\\
\and
Trung-Nghia Le\thanks{Corresponding author.}\inst{1,2}\orcidlink{0000-0002-7363-2610}}
%
\authorrunning{P.-T. Nguyen et al.}
\institute{University of Science, VNU-HCM, Ho Chi Minh City, Vietnam \and
Vietnam National University, Ho Chi Minh City, Vietnam\\
\email{\{21120028, 21120068\}@student.hcmus.edu.vn, \{tmtriet, ltnghia\}@fit.hcmus.edu.vn}}

\maketitle              


\begin{abstract}
This paper aims to bridge the semantic gap between visual content and natural language understanding by leveraging historical events in the real world as a source of knowledge for caption generation. We propose VisChronos, a novel framework that utilizes large language models and dense captioning models to identify and describe real-life events from a single input image. Our framework can automatically generate detailed and context-aware event descriptions, enhancing the descriptive quality and contextual relevance of generated captions to address the limitations of traditional methods in capturing contextual narratives. Furthermore, we introduce a new dataset, EventCap (https://zenodo.org/records/14004909), specifically constructed using the proposed framework, designed to enhance the model's ability to identify and understand complex events. The user study demonstrates the efficacy of our solution in generating accurate, coherent, and event-focused descriptions, paving the way for future research in event-centric image understanding.

\end{abstract}
\section{Introduction}
Image captioning aims to generate descriptive captions for images. However, most current methods tend to produce captions with a limited understanding of the image, focusing primarily on identifying objects, actions, and basic physical attributes \cite{krause2017hierarchical, wu2022grit, li2022mplug}. These approaches fall short in conveying deeper context, as they lack the ability to infer meaningful information about the events or interactions taking place in the image. This limitation becomes especially evident when the goal is to describe not just what is visible but also the underlying story or context associated with the image.

\begin{figure}[t!]
    \centering
    \includegraphics[width=1.0\linewidth]{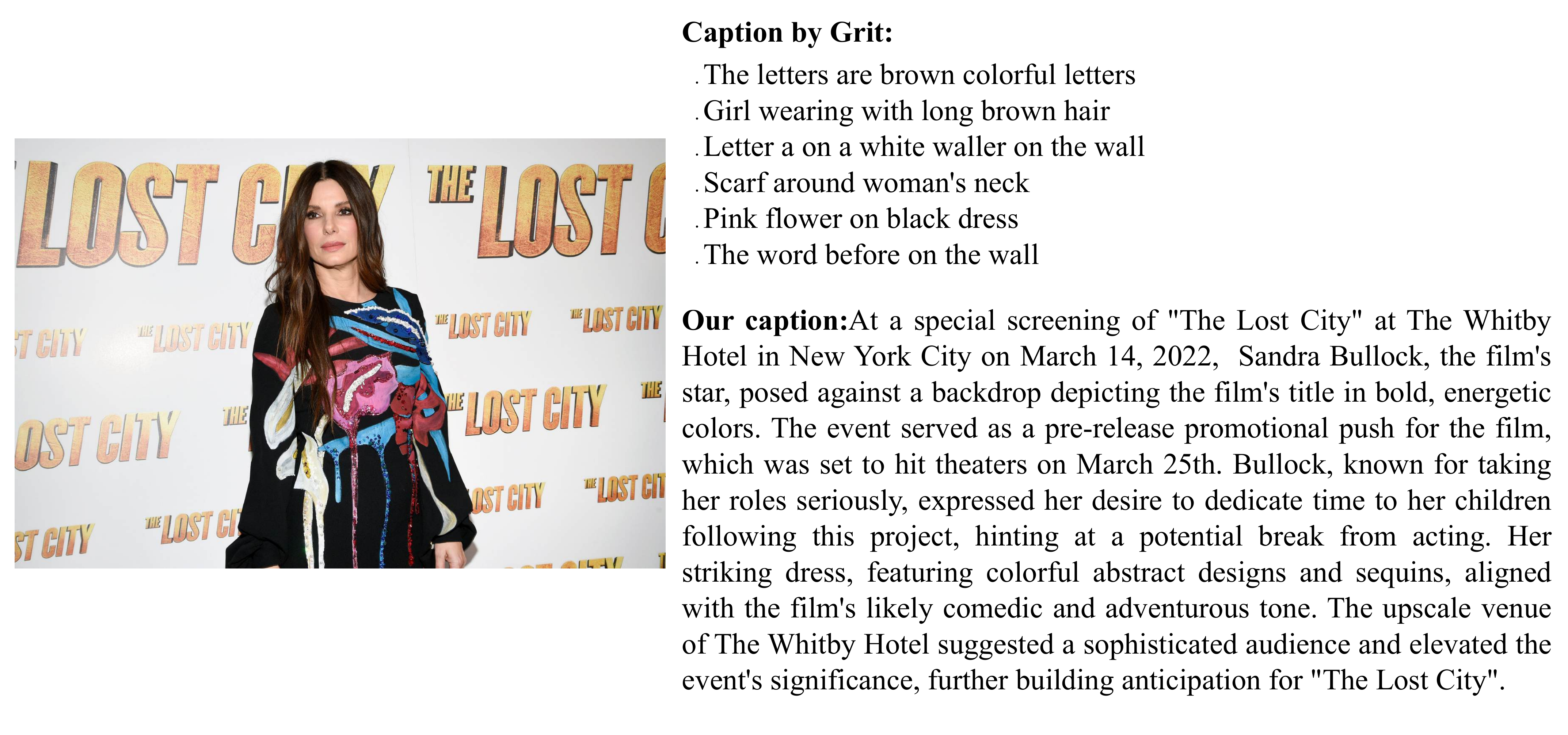}
    \caption{Comparison between general caption by Grit \cite{wu2022grit} and our event-based caption, highlighting additional details such as location, date, identity, and event purpose. 
    }
    \label{fig:teaser}
\end{figure}

In many cases, the generated captions are too superficial to capture complex scenarios where additional information, such as who is involved, what is happening, where and when the event took place, and its significance, is critical. As a result, these methods are inadequate for providing rich, informative captions that align with more sophisticated user needs, such as understanding or retrieving images related to real-world events. To overcome this, a new approach is needed—one that integrates contextual details and event-related information to create comprehensive, narrative-driven captions that go beyond simple object recognition.

Our research introduces the task of Event-Enriched Image Captioning (EEIC), which aims to generate captions that provide richer, more comprehensive information about an image. This approach is demonstrated through a sample depicted in Fig. \ref{fig:teaser}, where we showcase the result caption that our method generates. This example illustrates the enhanced descriptive quality and contextual depth that EEIC can bring to image captioning. 
These captions go beyond simple visual descriptions by offering deeper insights, including the names and attributes of objects, the timing, context, outcomes of events, and other crucial details—information that cannot be gleaned from merely observing the image. This approach facilitates the creation of more coherent and detailed narratives, capturing not only the visible elements but also the underlying context and significance of the scene, ultimately offering a more complete understanding of what the image represents. 

The core idea of our approach is to harness event-related information from credible sources while leveraging the reasoning capabilities of both vision-language models and large language models (LLMs). We propose a fully automated four-step framework, called VisChronos, to analyze both the visual content and the temporal, event-based aspects of the scene, ensuring a more comprehensive understanding. VisChronos operates through a systematic process designed to extract, analyze, and synthesize information from images and associated events. First, a vision-language model identifies and describes the most important aspects of the image, including both those specified by prompts and those deemed important by the model itself. Next, in the second step, an LLM generates questions about the image based on the key aspects identified in the first step, which includes both mandatory and optional questions. In the third step, another LLM answers these questions using event information that we provide. Finally, a separate LLM synthesizes and processes the information from the previous three steps to infer and generate the final caption for the image. Unlike traditional models that rely on learning from a massive dataset and risk generating information that may be fabricated or irrelevant, our method addresses this issue by incorporating factual sourcescredible, human-authored articles that provide real, context-rich information. By drawing directly from authentic articles and aligning this information with image content, our framework ensures that captions accurately represent real events in human history. This framework is designed to establish an efficient information mining flow, systematically dividing different stages of information extraction across each step. This approach ensures that the framework can mine the most useful and relevant information at each stage, ultimately resulting in rich and contextually accurate captions.


Extensive human evaluations of captions generated by the VisChronos reveal that they are comparable to captions crafted by human annotators. These machine-generated captions were particularly praised for their completeness, coherence, conciseness, and the inclusion of relevant information not explicitly visible in the images.

Using VisChronos, we have created a dataset named EventCap, consisting of 3140 event-based image-caption pairs. Each pair has been carefully curated using images and related information sourced from a wide range of credible articles. This collection serves as a valuable resource for training and evaluating the performance of image captioning models in understanding and describing complex real-world events. To the best of our knowledge, no similar dataset exists that is specifically designed for the task of event-enriched image captioning, making EventCap a unique and essential resource for advancing research in this area.

Our main contributions can be summarized as follows:
{
\begin{itemize}
    \item We introduce the new task of Event-Enriched Image Captioning (EEIC), which generates captions that provide profound insights into both the image and the events depicted, including information that cannot be inferred solely from the image.

    \item We propose a novel four-stage framework, namely VisChronos to solve the task of EEIC, combining vision-language models and LLMs to generate detailed, context-rich captions.

    \item We publicly release EventCap (https://zenodo.org/records/14004909), a dataset consisting of 3140 image-caption pairs, processed from 491 CNN articles.

\end{itemize}
}
 
\section{Related Work}
\subsection{Dense captioning}
Dense captioning, which aims to generate detailed descriptions for objects in scenes or videos, remains a challenging task. Over the years, several methods have been developed to address this problem, each employing distinct approaches and making notable contributions. Chen et al. \cite{chen2020scan2cap} introduced Scan2Cap, an end-to-end method for dense captioning in RGB-D scans, utilizing 3D point cloud inputs to generate bounding boxes and corresponding object descriptions. Building on this, Wang et al. \cite{wang2021dense} presented PDVC, a framework that formulates dense video captioning as a task of set prediction, enabling efficient parallel decoding. Aafaq et al. \cite{aafaq2022dense} proposed the ViSE framework and VSJM-Net, which leverage early linguistic information fusion to model word-context distributional properties for improved dense video captioning. Shao et al. \cite{shao2023textual} further advanced the field by incorporating textual context-aware methods that generate diverse and context-rich captions. Similarly, Jiao et al. \cite{jiao2022more} presented MORE, a model that captures complex scene relations for superior captioning accuracy . Furthermore, Wu et al.\cite{wu2022grit} developed GRiT, a generative region-to-text transformer model that emphasizes object understanding in dense captioning tasks.



Despite the advancements made by these approaches, they typically generate conventional captions that lack real-world semantic depth, as they rely solely on information from the image or video. Moreover, most existing methods produce captions in a single pass through learned representations, which can result in missing critical details. In contrast, our method continuously supplements the captioning process through an interactive dialogue between models, allowing for the extraction of more nuanced and semantically rich information.



\subsection{Large Language Models}
Large Language Models (LLMs) have garnered significant attention due to their ability to generate human-like text and solve complex tasks across various domains. 
%
%
GPT (Generative Pre-trained Transformer), developed by OpenAI, is one of the most well-known LLMs. Floridi et al. \cite{floridi2020gpt3} introduced GPT-3, a third-generation autoregressive language model that generates human-like text using deep learning techniques. They explored its nature, scope, limitations, and potential consequences, emphasizing that GPT-3 is not intended to pass complex mathematical, semantic, or ethical tests . Brown et al. \cite{brown2020fewshot} further demonstrated that scaling language models, such as GPT-3 with 175 billion parameters, significantly improves few-shot performance across various tasks, sometimes outperforming state-of-the-art (SOTA) approaches. 
%
In subsequent developments, Achiam et al. \cite{achiam2023gpt4} presented GPT-4, a large-scale multimodal model capable of processing both image and text inputs to generate text outputs, marking a significant advancement in multimodal language modeling. 
Meanwhile, Yang et al. \cite{yang2023lmm} explored GPT-4V, expanding GPT-4's capabilities to include vision-based tasks, opening new avenues for large multimodal models. In addition, Wang et al. \cite{wang2023decodingtrust} evaluated the trustworthiness of GPT models, concluding that GPT-4 is generally more reliable than GPT-3.5 but remains vulnerable to adversarial attacks, such as jailbreaking or misleading prompts.

Gemma is a lightweight, SOTA open model that builds upon the research and technology developed for Gemini models. As outlined in recent studies, Gemma has exhibited strong performance across various benchmarks for language understanding, reasoning, and safety \cite{gemma2024}. Notably, Gemma has been rigorously evaluated alongside other large language models (LLMs) to assess its capabilities across different languages, modalities, models, and tasks \cite{ahuja2023megaverse}. The advancements in Gemma are a result of ongoing improvements in multimodal research, with enhancements in understanding and processing non-textual data inputs such as images and speech, which significantly contribute to its versatility in both academic and practical applications. Furthermore, by being available as an open-source model, Gemma provides accessible, cutting-edge AI technology to the broader research community, while also offering paid solutions for advanced functionalities and commercial deployment \cite{gemma2024}.

In this framework, we integrate the use of dense captioning models with both paid and open-source LLMs, including models such as GPT and Gemma, to leverage their combined strengths for enhanced performance across a range of tasks. By this framework, we also generate the EventCap dataset, the first dataset specifically designed for event captioning, providing a unique resource for accurately describing event contexts in diverse applications.


\section{Proposed VisChronos framework}

\begin{figure}[t!]
\includegraphics[width=\textwidth]{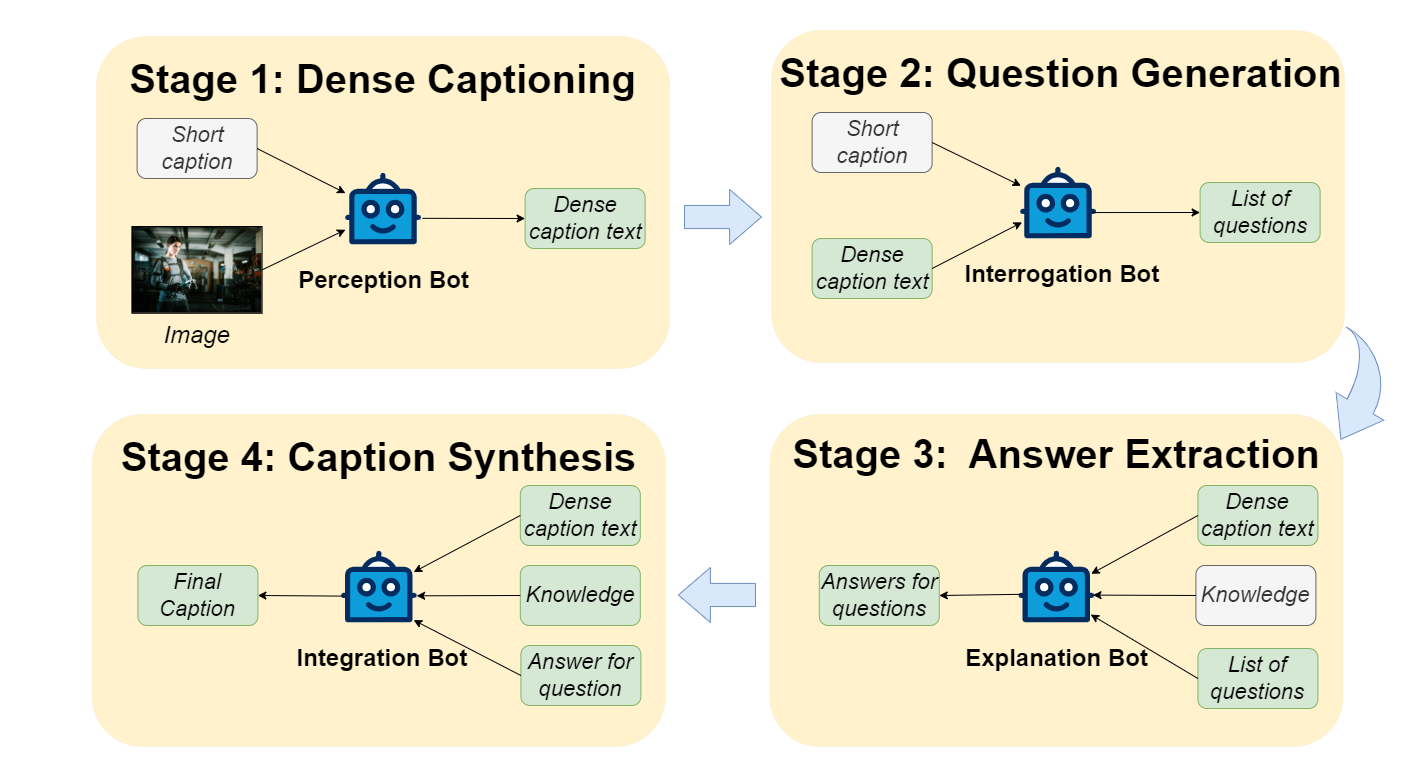}
\caption{VisChronos framework for Event-Enriched Image Captioning.} \label{frameworkfig}
\end{figure}

\subsection{Overview}

In this work, we introduce a novel multi-stage framework aimed at enhancing image captioning by generating contextually rich captions that go beyond describing the image. Our approach captures not only visual elements but also events, facts, and broader contextual information inferred from the image. As shown in Fig. \ref{frameworkfig}, our framework consists of four stages, each of which is handled by a dedicated bot, designed to perform a specific task and feed information to the next stage. These stages are:

\begin{itemize} 
\item 
\textbf{Stage 1 (Dense Captioning - Perception Bot)}: Generates a detailed description of the image, capturing objects, people, actions, and contextual elements. 
\item \textbf{Stage 2 (Question Generation - Interrogation Bot)}: Creates a set of questions based on the image description to explore the event's context and details. 
\item \textbf{Stage 3 (Answer Extraction - Explanation Bot)}: Extracts answers to the questions from external knowledge sources, ensuring that answers are provided only when the model is confident. 
\item \textbf{Stage 4 (Caption Synthesis - Integration Bot)}: Synthesizes the dense caption, answers, and external knowledge into a comprehensive and contextually enriched final caption. 
\end{itemize}

Each stage is designed to progressively enrich the information related to the image, ultimately producing a caption that is both visually descriptive and contextually informative.

\subsection{Stage 1: Dense Captioning - Perception Bot}
In this stage, the Perception Bot is responsible for generating a dense caption that thoroughly describes the image’s contents, including objects, people, actions, and background elements. This process is guided by a carefully crafted instruction provided to dense captioning model. The instruction ensures that the caption captures all significant visual elements, with attention to detail, relationships between objects, and overall context. Therefore, the dense caption can serve as the foundation for the entire framework, offering a detailed and structured description of the scene. This stage results in a rich description of the image with detailed information of objects, actions, and relationships, that serves as input for subsequent stages.







\subsection{Stage 2: Question Generation - Interrogation Bot}

The second stage is crucial for extracting the deeper context and events surrounding the image. This stage focuses on generating a set of structured questions that aim to explore various aspects of the event or scene depicted in the image. The questions are designed to extract more information from the accompanying article or external sources, leading to a comprehensive understanding of the depicted event.

 
A structured approach is employed to generate questions that comprehensively address the event or context depicted in the image. The questions are crafted to ensure that all key aspects of the scenario are explored, providing a robust foundation for the subsequent explanation and synthesis stages as shown in Table \ref{stage2table}.

\begin{table}[t!]
\centering
\caption{Key aspects covered by the questions in Question Generation stage.}
\label{stage2table}
\begin{tabular}{|l|p{8cm}|}
\hline
\textbf{Category} & \textbf{Description} \\ 
\hline
Time & Questions addressing when the event occurred or its timeline to understand the temporal context. \\ 
\hline
Context or Reason & Questions exploring the circumstances or motivations behind the event, seeking to clarify why the event took place. \\ 
\hline
Main Events & Questions focusing on the central actions or occurrences within the event, ensuring a clear understanding of the primary narrative. \\ 
\hline
Outcome & Questions that inquires about the result or conclusion of the event, aiming to highlight the final impact or resolution. \\ 
\hline
Impact & Questions exploring the broader consequences or effects of the event on individuals, groups, or larger contexts. \\ 
\hline
Objects or People & Several questions delving into the key people or objects mentioned in the dense caption, ensuring that all significant elements are covered. \\ 
\hline
Special Figures & Specific questions about notable or important figures involved in the event, shedding light on their roles and influence. \\ 
\hline
Emotions and Reactions & Questions designed to explore the emotional states or reactions of the people in the image, providing insight into the human element of the scene. \\ 
\hline
Background Details & Questions addressing the setting or background elements in the image, helping to paint a fuller picture of the environment in which the event takes place. \\ 
\hline
Future Implications & Questions speculating on the potential future outcomes or ramifications of the event, aiming to place the event within a broader temporal and societal context. \\ 
\hline
\end{tabular}
\end{table}

The design of the question-generation model ensures that each of these dimensions is adequately explored, leading to a comprehensive inquiry into the event depicted in the image. This stage serves as the foundation for the next phase, where the generated questions are used to retrieve detailed answers from external knowledge sources.


\subsection{Stage 3: Answer Extraction - Explanation Bot}

The third stage focuses on extracting detailed answers to the questions generated in Stage 2 by leveraging external knowledge sources such as articles or accompanying text related to the image. The Explanation Bot ensures that the answers are accurate and directly relevant to the event depicted in the image.


This stage requires a highly precise approach to ensure that the model provides accurate answers grounded in the available information. A key principle guiding the model’s behavior during this phase is certainty. The model is instructed to answer questions only when it is 100\% confident in the information it provides. This ensures that the answers are factually reliable and directly linked to the knowledge from the article or dense caption.

\textbf{Certainty Rule:} If the model is unable to locate relevant information or if it is unsure about the correctness of an answer, it must respond with "no information." This instruction prevents the model from speculating or providing potentially misleading or inaccurate answers.

\textbf{Information Retrieval:} The model focuses on extracting factual data from the article or other external knowledge sources and cross-referencing this information with the dense caption to ensure consistency and accuracy. The goal is to answer each question as fully as possible, but only when the necessary information is available and can be confidently inferred.

This strict adherence to certainty ensures that the answers provided are reliable and fact-based, maintaining the integrity of the image-captioning process. By instructing the model to explicitly state "no information" when necessary, the framework avoids overgeneralization or the inclusion of speculative answers.


\subsection{Stage 4: Caption Synthesis - Integration Bot}

The final stage synthesizes the information gathered from previous stages into a comprehensive caption that reflects both the visual content of the image and the broader contextual information.


The instruction provided to the model for this stage ensures that it combines the dense caption, external knowledge, and question responses to generate a detailed and contextually enriched final caption. This synthesis produces a narrative that reflects both the visual and factual elements of the image. The final caption aims to describe the event depicted in the image, taking into account the knowledge gathered in the previous stages.

Output of this stage is a final caption that provides a rich, event-based description of the image, integrating visual details with contextual information from external sources.

\subsection{Detailed Implementation}

Our VisChronos framework is designed with flexibility, allowing  integration with a variety of dense captioning models and LLMs, such as GRIT, GPT, Gemma, Bard, etc. This adaptability ensures that different models can be used for various tasks within the framework. However, for the creation of the EventCap dataset, we specifically employed GPT-4o as the vision-language model for dense captioning in the first step, and utilized the Gemma-2-9b-it (open-source version) model for stages 2, 3, and 4 to handle question generation, answer extraction, and final caption synthesis.

\section{Proposed EventCap Dataset}


\subsection{Dataset Construction}

To build the EventCap dataset using our VisChronos framework, we required both images of real-world events and the corresponding event-related information, which we refer to as "knowledge." Specifically, we crawled data from 491 articles published by CNN between 2014 and 2022, covering a wide range of categories, including \textit{business, entertainment, health, news, politics}, and \textit{sport}. For each article, we collected the full textual content and all images included within the article.

\begin{figure}[t!]
    \centering
    \makebox[\textwidth]{\includegraphics[width=1.0\linewidth]{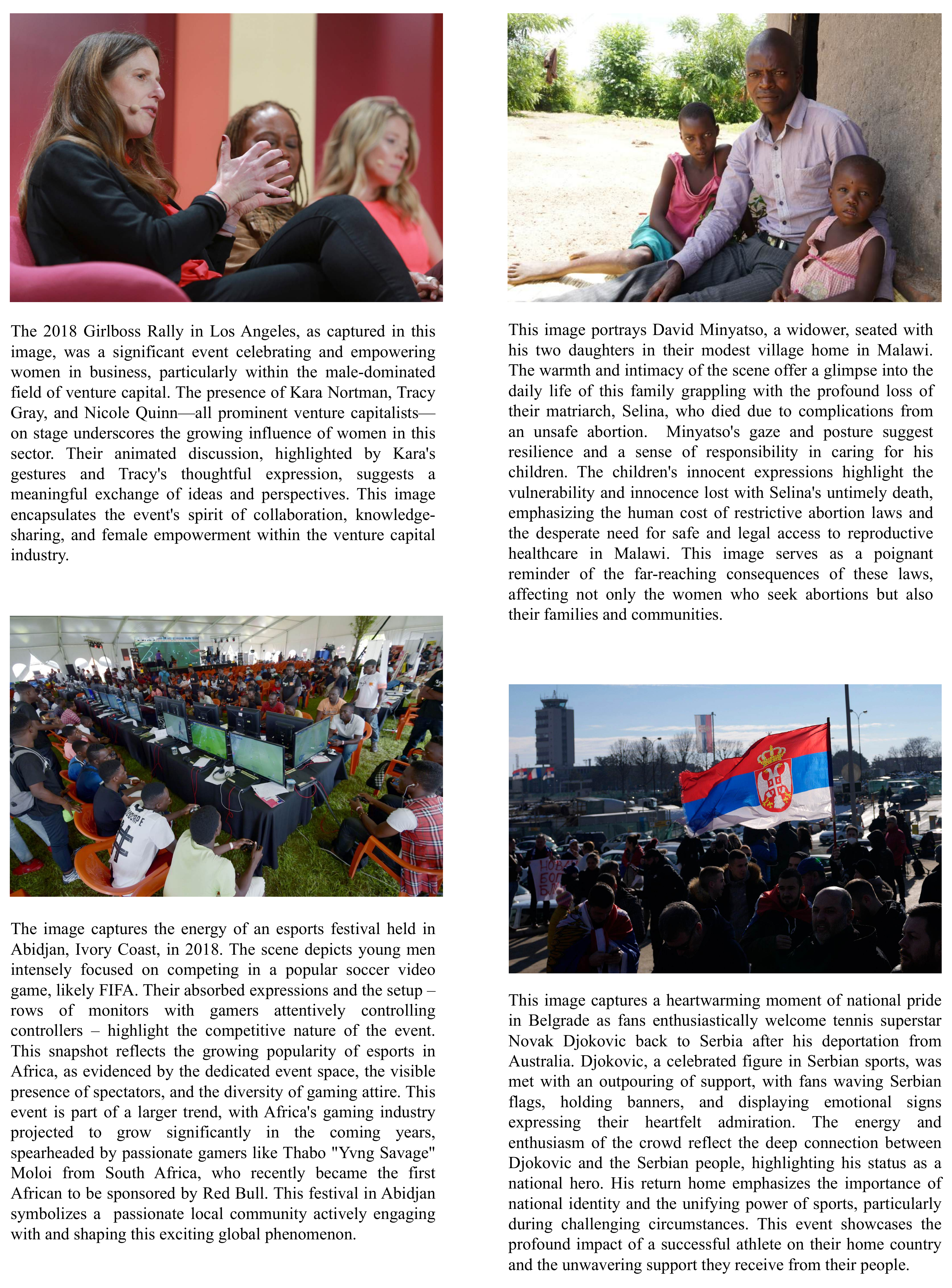}}
    \caption{Examples of image-caption pairs generated by VisChronos.}
    \label{fig:example}
\end{figure}

The dataset creation process involved processing each image along with the corresponding article content, which served as external knowledge, through the VisChronos framework. The images were sourced from reputable news sites and captured by human photographers, ensuring authenticity and real-world relevance. The framework then generated captions for each image, using the article’s content to ensure contextually relevant and detailed event descriptions. Captions generated by our method, as evaluated in Section \ref{sec:evaluation}, achieved a quality level comparable to human-written descriptions. This procedure formed the foundation for creating each image-caption pair in the dataset. Examples of image-caption pairs generated by our framework are illustrated in Fig. \ref{fig:example}.

\subsection{Dataset Specifications and Statistics}

\begin{figure}[t!]
    \centering
    \includegraphics[width=0.75\textwidth]{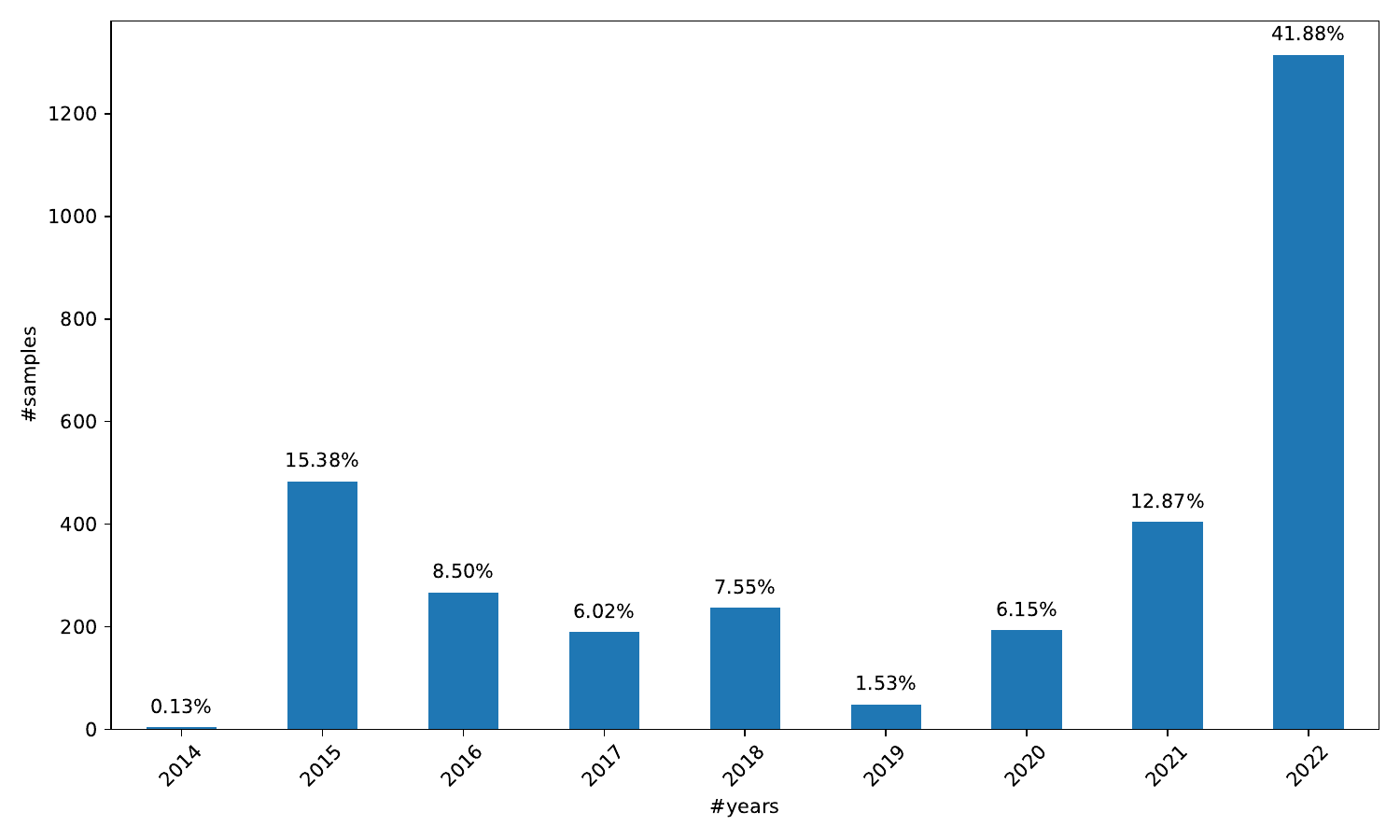}
    \caption{Sample distribution in EventCap dataset by year (best view in color \& zoom-in).}
    \label{fig:images_by_year_percentage}
\end{figure}

\begin{figure}[t!]
    \centering
    \includegraphics[width=0.55\textwidth]{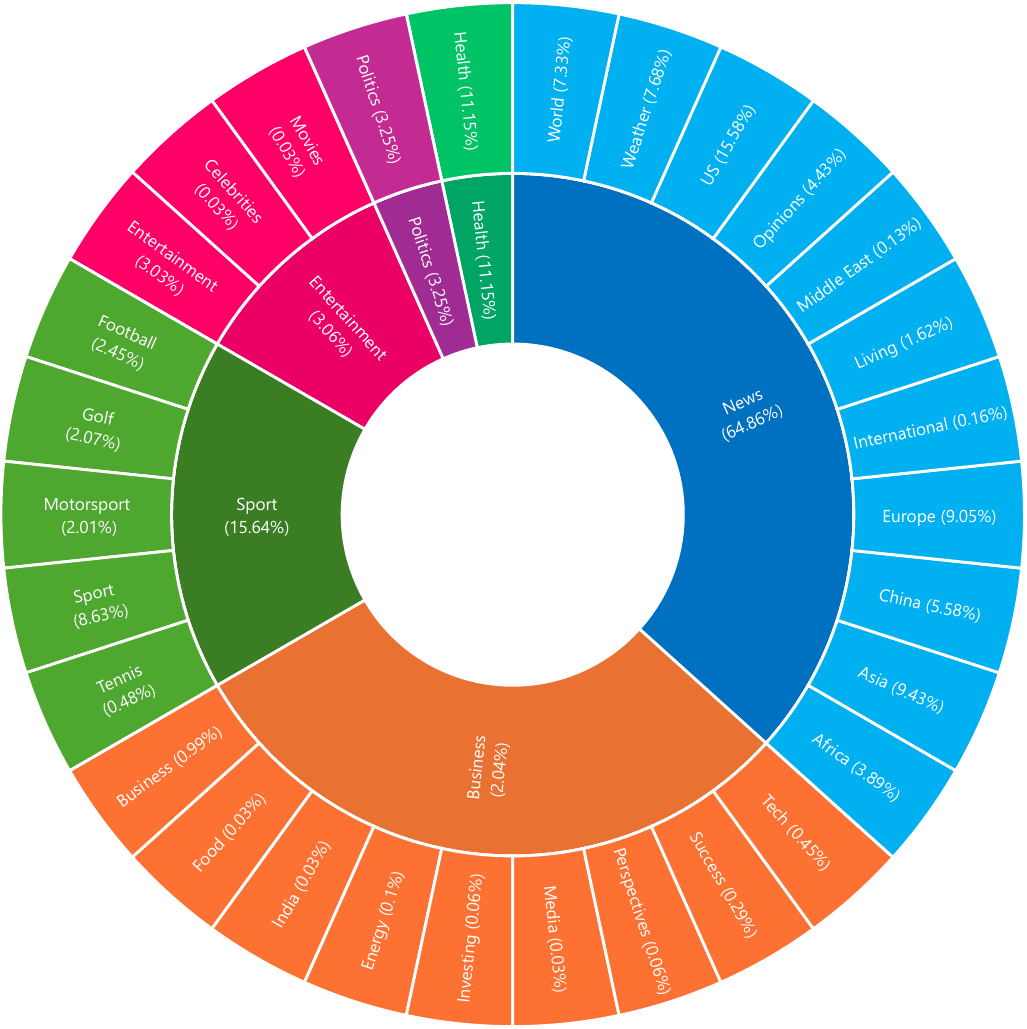}
    \caption{Sample distribution in EventCap dataset by category and section (best view in color \& zoom-in).}
    \label{fig:sunburst_dataset}
\end{figure}

The EventCap dataset comprises 3140 image-caption pairs, with each pair considered a single sample, specifically curated to enhance the understanding of complex, event-driven content. The samples in the EventCap dataset are distributed across the years 2014 to 2022, covering a wide range of categories, including \textit{business, entertainment, health, news, politics}, and \textit{sport}; each categories having different sections. The images prominently feature current events, capturing significant moments in these years. These images vary widely in size, from 532 to 5440 pixels in width and 338 to 4618 pixels in height. Captions are generated using our VisChronos framework, designed to distill and articulate the essence and nuances of each event in a narrative style. These captions are extensive, ranging from 50 to 793 words with an average length of 106 words, and are structured in one or more paragraphs to convey detailed, context-rich descriptions of the depicted scenes. Fig. \ref{fig:images_by_year_percentage} presents sample distribution by years in the EventCap dataset while Fig. \ref{fig:sunburst_dataset} breaks down the sample distribution by categories, illustrating the dataset’s diversity.





\subsection{User Study}
\label{sec:evaluation}

Since this task is novel, the existing metrics typically used to evaluate traditional models are not directly applicable. Furthermore, as this is the first dataset specifically created for this task, there is no comparable dataset available for direct benchmarking. To address this, we conducted a user study aimed at assessing both the effectiveness of our method and the quality of the generated dataset. 
Additionally, the study aimed to compare the quality of captions generated by our framework with those written by humans, highlighting the strengths and weaknesses of each approach.

\paragraph{Participants: } We invited 10 participants (8 males, 2 females, aged 18 to 30) to participate in our study. Participants were selected from diverse backgrounds, including professionals in media, journalism, and computer science, to ensure a comprehensive and well-rounded evaluation.


\paragraph{Apparatus and procedure:}

Our study was conducted both online and on-site in our lab, where participants completed the tasks. Each participant received clear instructions on evaluating captions and writing their own for comparison. They were required to spend at least 4 minutes evaluating and 5 minutes writing the caption for each image including reading the corresponding article. The total time for the study sessions was approximately 120 minutes per participant.

\begin{figure}[t!]
    \centering
    \includegraphics[width=0.8\linewidth]{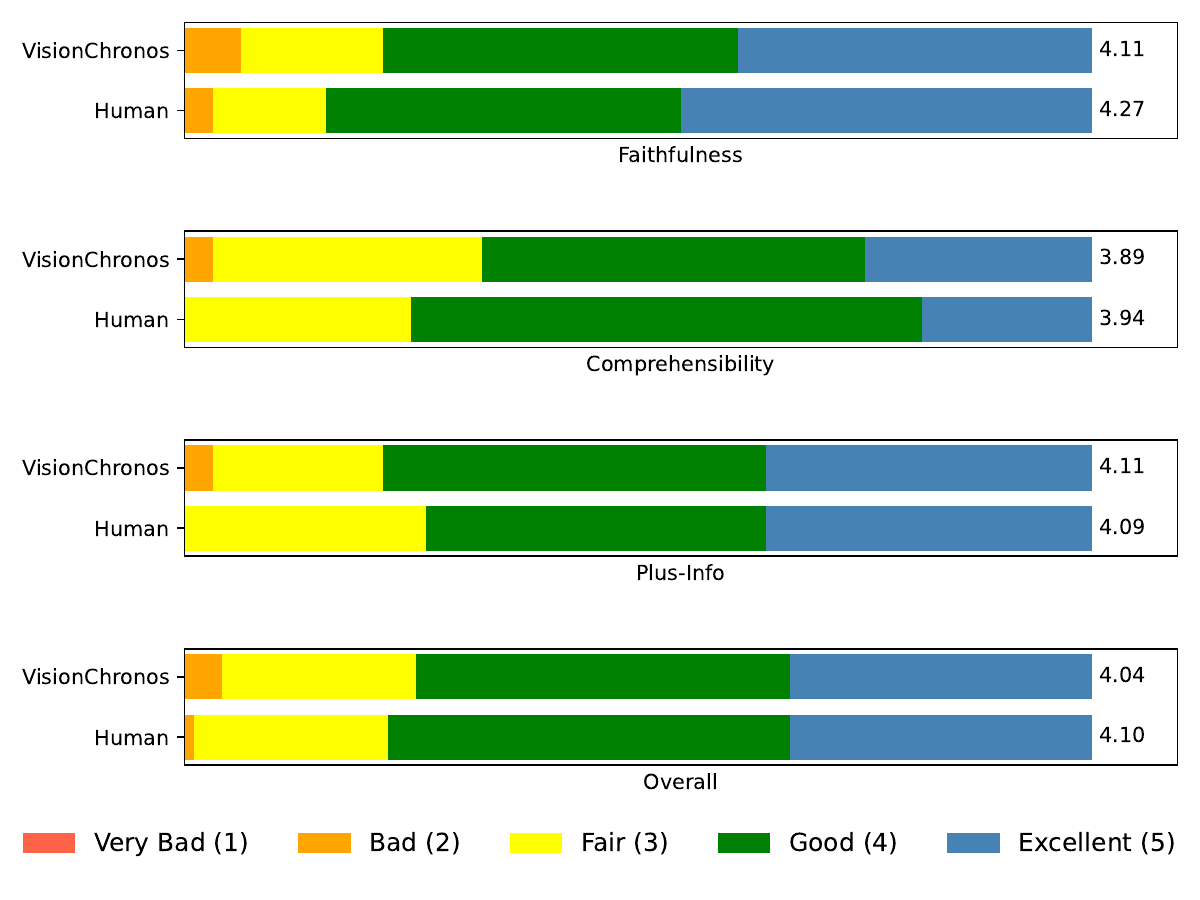}
    \caption{Comparative performance of human and VisChronos in writing event-based image captions across metrics. Our proposed method can achieve human-level writing (best view in color \& zoom-in).}
    \label{fig:user_study}
\end{figure}

First, participants were asked to write their own captions for 5-10 images from 2 articles. After that, participants evaluated the quality of captions for approximately 10-15 images from 4 different articles. Half of the captions were generated by our framework, while the other half were written by humans (i.e., other participants).

The participants were asked to rate the performance of the captions on a scale of 1 to 5 across three metrics, based on their individual perspectives. The comparison was based on several key metrics:

\begin{itemize} 
    \item \textbf{Faithfulness}: Whether the content of the caption fully describes the key events depicted in the image. 
    \item \textbf{Comprehensibility}: Whether the caption is concise, easy to read, and free of unnecessary information. 
    \item \textbf{Plus-Info}: Whether the caption provides additional useful information that cannot be inferred directly from the image. 
\end{itemize}

\paragraph{Quantitative results}
We obtained quantitative results by averaging the ratings of each participant across three metrics and then averaging those results over all participants. These outcomes shown in Fig. \ref{fig:user_study} indicate that the performance of the VisChronos framework is not significantly different from that of human-written captions across evaluated metrics. This highlights potentials of our solution in real-life applications such as writing image description for newspapers, journals, books, etc. The user study results also demonstrates the high quality of our EventCap dataset in the development of event-based image captioning models.


\paragraph{Limitations}
Our method relies heavily on the quality of the accompanying information to produce accurate captions. Incomplete or poor-quality data can negatively impact the results. Additionally, the four-step process increases the time required, with an average of 1 minute needed to caption each image, which may limit scalability for large datasets.

\comment{

}

\section{Conclusion}
In this paper, we introduced the VisChronos framework, a novel four-stage approach designed to tackle the new task of Event-Enriched Image Captioning (EEIC). This task aims to generate captions that not only describe the visual content of an image but also provide insights into the underlying events, incorporating information beyond what is directly observable. By leveraging a combination of vision-language models and large language models (LLMs), our framework iteratively refines the captioning process through interactive dialogue between models, resulting in semantically richer and contextually enhanced captions. Additionally, we presented EventCap dataset, consisting of 3140 image-caption pairs from 491 CNN articles, which serves as a strong foundation for future research in event-based image captioning and related tasks. To the best of our knowledge, EventCap is the first dataset specifically created for the task of event-enriched image captioning, establishing it as a pioneering resource and a strong foundation for future research in event-based image captioning and related tasks.

As part of ongoing efforts, we aim to significantly expand the EventCap dataset to 20,000 articles and 50,000 images, creating a comprehensive resource for the research community to advance event-enriched image captioning and multimodal learning. Additionally, we plan to refine the VisChronos framework into a multimedia tool that enhances human experiences by integrating visual, textual, and contextual information. This evolution will support applications like event detection, real-time image analysis, and interactive storytelling, leveraging the framework’s ability to generate rich, event-driven captions.

\section*{Acknowledgement } 
This research is supported by research funding from Faculty of Information Technology, University of Science, Vietnam National University - Ho Chi Minh City.

%
%
%
%

\bibliographystyle{splncs04}
\bibliography{ref/References}
\end{document}